\documentclass[10pt,twocolumn,letterpaper]{article}

\usepackage{iccv}
\usepackage{times}
\usepackage{epsfig}
\usepackage{graphicx}
\usepackage{amsmath}
\usepackage{amssymb}
\usepackage{authblk}
\newcommand*{\affaddr}[1]{#1}
\newcommand*{\affmark}[1][*]{\textsuperscript{#1}}
\newcommand*{\email}[1]{\texttt{#1}}
\usepackage{url}

\usepackage[breaklinks=true,bookmarks=false]{hyperref}

\iccvfinalcopy 

\def\iccvPaperID{426} 

\ificcvfinal\fi

\begin{document}

\title{Human Mesh Recovery from Monocular Images\\ via a Skeleton-disentangled Representation}

\author[ ]{Yu Sun\affmark[1]\thanks{This work is done when Yu Sun is an intern at JD AI Research.}}
\author[ $\star$]{Yun Ye\affmark[2]}
\author[ $\star$]{Wu Liu\affmark[2]\thanks{Corresponding author: Wu Liu. $\star$ Equal contribution.}}
\author[ ]{Wenpeng Gao\affmark[1]}
\author[ ]{YiLi Fu\affmark[1]}
\author[ ]{Tao Mei\affmark[2]}
\affil[ ]{\affaddr{\affmark[1]Harbin Institute of Technology    }\affaddr{\affmark[2]JD AI Research}}
\affil[ ]{\small \texttt{yusun@stu.hit.edu.cn  yun.ye@intel.com  liuwu@live.cn}}
\affil[ ]{\small \texttt{\{wpgao, meylfu\}@hit.edu.cn tmei@live.com}}

\maketitle
\ificcvfinal\thispagestyle{empty}\fi

\begin{abstract}
We describe an end-to-end method for recovering 3D human body mesh from single images and monocular videos. Different from the existing methods try to obtain all the complex 3D pose, shape, and camera parameters from one coupling feature, we propose a skeleton-disentangling based framework, which divides this task into multi-level spatial and temporal granularity in a decoupling manner. In spatial, we propose an effective and pluggable ``disentangling the skeleton from the details'' (DSD) module. It reduces the complexity and decouples the skeleton, which lays a good foundation for temporal modeling. In temporal, the self-attention based temporal convolution network is proposed to efficiently exploit the short and long-term temporal cues. Furthermore, an unsupervised adversarial training strategy, temporal shuffles and order recovery, is designed to promote the learning of motion dynamics. The proposed method outperforms the state-of-the-art 3D human mesh recovery methods by 15.4\% MPJPE and 23.8\% PA-MPJPE on Human3.6M. State-of-the-art results are also achieved on the 3D pose in the wild (3DPW) dataset without any fine-tuning. Especially, ablation studies demonstrate that skeleton-disentangled representation is crucial for better temporal modeling and generalization. The code is released at \color{blue}\small\url{ https://github.com/Arthur151/DSD-SATN}.

\end{abstract}
\section{Introduction}

Different from traditional 3D pose estimation that usually predicts the location of 14/17 skeleton joints~\cite{tpnet,simple}, 3D human body mesh recovery from the monocular images is a more complex task, which tries to estimate the more detailed 3D shape and joint angles. In detail, it needs to estimate more than 85 parameters, which controls 6890 vertices~\cite{smpl} that form the surface of 3D body mesh. Moreover, the information loss from the 3D scene to a 2D image, inherent ambiguity, and complex changes in human body shape and pose further increase the complexity of this task. Therefore, although the 3D body mesh recovery is important in computer vision, motion/event analysis~\cite{liu2015devnet,liu2016recognizing,liu2013accurate,liu2017progressive,liu2018t}, and virtual try-on~\cite{dong2019flow,dong2019mgvton}, it is still a frontier challenge. In this paper, we try to solve this problem via a skeleton-disentangled representation in multi-level spatial and temporal granularity.

\begin{figure}
	\begin{center}
		\includegraphics[width=1.00\linewidth]{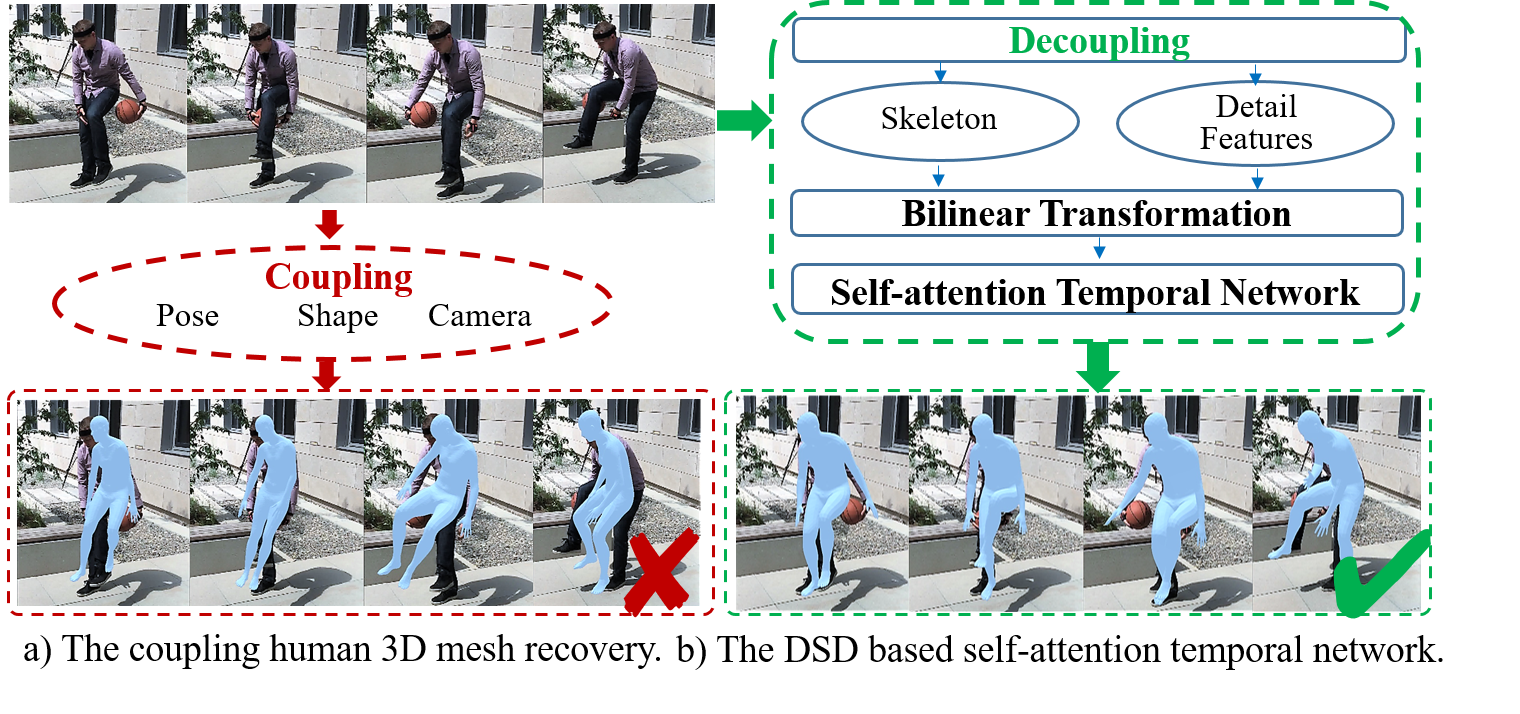}
		\caption{Human 3D mesh recovery from monocular video based on skeleton disentangling and temporal coherence.}
		\label{fig:videoresult}
	\end{center}
\vspace{-6mm}
\end{figure}

Most of the existing methods try to recover the human 3D mesh from single images. Previous multi-stage approaches~\cite{unite,neuralbodyfitting,humanshape} first extract human body information (e.g. 2D pose and segmentation) and then estimate the 3D model parameters from them, which is typically not optimal. Recently, HMR~\cite{hmr} provides an end-to-end solution to learn a mapping from image pixels directly to model parameters, which shows a significant performance advantage over the two-stage methods. However, as shown in Figure~\ref{fig:videoresult}, severe coupling problem makes the prediction unstable. In HMR, 3D pose, body shape, and camera parameters are derived directly from the same feature vector using the fully connected layers. Without any decoupling measurements, high complexity makes features of different targets tightly coupled. Moreover, existing datasets with 3D annotations are collected in a constrained environment with limited motion and shape patterns. Therefore, many previous methods have employed 2D in-the-wild pose datasets, like MPII~\cite{mpii}, to learn richer poses. It may be sufficient for 3D pose estimation. While, for 3D mesh recovery, 2D pose is far from enough for recovering the complex human 3D shape and pose details. Lacking supervision makes the predictions of the details vulnerable, which further exacerbates the coupling problem. The coupling strategy makes the model trained on these datasets cannot well generalize to the complex environments and various human states.

To solve this problem, we propose a lightweight and pluggable DSD module to decouple different factors in an end-to-end manner. The main idea of DSD module is to disentangle the skeleton from the 3D mesh details with bilinear transformation. Firstly, information of the pose skeleton and the rest details (e.g., body shape, detailed pose information) are extracted independently. Furthermore, the bilinear transformation is employed to aggregate two pieces of information while keeping their decoupling in the new feature space. Finally, the network is trained end-to-end to keep the global optimal. In the evaluations, we demonstrate that the DSD module outperforms the state-of-the-art methods~\cite{kanazawa2018learning} by 13\% PA-MPJPE on Human3.6M dataset. Moreover, the evaluations also demonstrate that the proposed portable DSD module can be easily plugged into other 2D/3D pose estimation network for recovering 3D human mesh.

Recovering human 3D mesh from single images may suffer from the inherent ambiguity as multiple 3D poses and shapes can be mapped to the same situation in a 2D image. To tackle this problem, we propose a self-attention temporal network (SATN) for efficiently optimizing the coherence between predictions of adjacent frames. SATN is the combination of self-attention module and TCN. TCN is employed for its efficient parallel computation and excellent short-term modeling ability. However, the inherently hierarchical structure of convolution layers limits the representation learning of the long-term sequence. More specifically, if the distance of two frames is larger than the kernel size of a single convolution layer, the model requires a stack of convolution layers to construct a long-term connection to relate them. As a result, the associations of the entire sequence cannot be established until the upper layer, which is inefficient. Therefore, we need a more efficient network to establish associations as early as possible. Based on this thought, self-attention is employed to associate the temporal features before TCN. Valuable connections among all frames can be established within just one self-attention layer. In this manner, associations can be efficiently established at the bottom of the temporal network, which greatly helps TCN  efficiently learning the short and long-term temporal coherence.

Given a video clip centered at frame $t$, SATN is developed to estimate human 3D mesh of frame $t$. Correspondingly, the supervision is only for the frame $t$. It is hard to determine whether SATN has learned the long-term temporal correlation. In other words, we lack supervision for guiding the temporal  representation learning. Therefore, we propose an unsupervised adversarial training strategy for learning the temporal correlation from the order of video frames. In detail, frames of a video clip are first shuffled and then re-sorted using the self-attention and sequence sorting module. By recovering the correct temporal order of the motion sequence in the video, a strong supervision signal is generated for learning the motion dynamics. Besides, considering that the motion orders are reversible and the adjacent poses are similar, the target of sequence sorting is well designed to meet these properties.

Compared with the DSD network, we report an additional 4.2\%/7.3\% improvements brought by our temporal model, in terms of PA-MPJPE on Human3.6M and 3DPW respectively. The proposed approach outperform the state-of-the-art methods~\cite{kanazawa2018learning,yoshiyasu2018skeleton} that predict 3D mesh by 15.4\% MPJPE and 23.8\% PA-MPJPE on Human3.6M. Besides, state-of-the-art results are also achieved on 3DPW without any fine-tuning. Especially, using features reorganized by DSD module for temporal modeling relatively improves 11.1\% and 27.5\% PA-MPJPE on 3DPW and Human3.6M respectively. It demonstrates that skeleton-disentangled representation is critical for better temporal motion modeling and generalization.

In summary, the following contributions are made:
\begin{enumerate}
\item An effective and portable DSD module is proposed to disentangle the skeleton from the rest part of the human 3D mesh. It reduces the complexity and lays a good foundation for better temporal modeling and ge.
\item Self-attention temporal network is proposed to learn the short and long-term temporal coherence of 3D human body efficiently. 
\item An unsupervised adversarial training strategy is proposed to guide the representation learning of motion dynamics in the video. 
\item Our entire framework is trained in an end-to-end manner. We outperform previous approaches that output 3D meshes~\cite{hmr,kanazawa2018learning,humanshape,yoshiyasu2018skeleton} in terms of 3D joint error.
\end{enumerate}

\label{sec:intro}
 \begin{figure*}
	\begin{center}
		\includegraphics[width=1\linewidth]{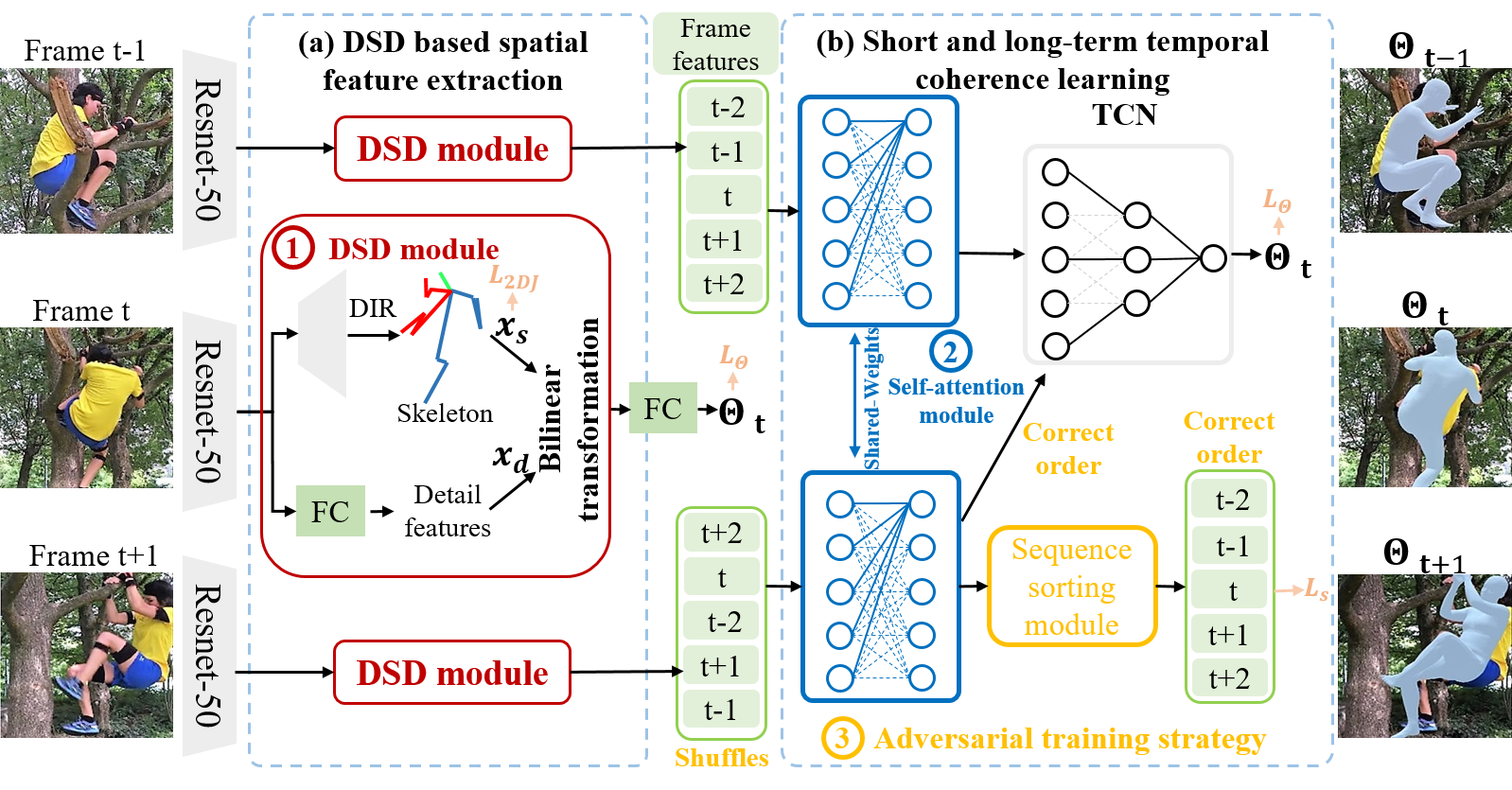}
		\caption{Overview of the proposed skeleton-disentangling based self-attention temporal network.}
		\label{fig:flowchart}
	\end{center}
\end{figure*}

\section{Related Work}
\label{sec:related work}
\textbf{Recovering 3D human mesh from single images.}
Most of the existing approaches formulated this problem as recovering the parameters of a statistical human body model, SMPL~\cite{smpl}. Recently proposed ConvNet-based methods~\cite{hmr,neuralbodyfitting,self,humanshape,yoshiyasu2018skeleton,unite,monocap,monoperfcap} have shown impressive performance. They can be split into two categories: two stages, direct estimation. 

The two-stage methods~\cite{neuralbodyfitting,humanshape,unite} first predict intermediate results, like human parsing~\cite{neuralbodyfitting}, and then predict the SMPL parameters from them. For instance, Pavlakos \etal \cite{humanshape} developed two individual networks to infer pose and shape from silhouettes and keypoint locations separately. The two-stage methods are robust to the domain shift, but throw away the important details of human body.

Some other methods~\cite{hmr,self} directly estimate SMPL parameters from images in an end-to-end manner. In particular, HMR~\cite{hmr} designed a discriminator to distinguish the authenticity of predicted SMPL parameters. In this generative adversarial manner, HMR could be trained with only 2D pose annotations. HMR greatly outperformed all two-stage methods. It may suggest that the end-to-end and non-disruptive feature extraction process is important in this task. Besides, Tung~\etal~\cite{self} developed various supervision approaches, like 2D re-projection of joints and segmentation, to better utilize all available annotations. However, as we introduced before, these methods suffer from severe coupling problem.

To tackle the coupling problem, we propose a DSD module to disentangle the skeleton from the rest details. It reduces network complexity and improves the accuracy of 3D pose recovery. In this process, the detailed information of 2D images is well preserved. The proposed DSD module based single-frame network outperforms all these methods. 

\textbf{Recovering 3D pose from monocular video.} 
We focus on 3D pose estimation from monocular video using deep networks, which is the most similar to ours. Various networks are developed to exploit temporal information. For example, TP-Net~\cite{tpnet} and Martinez \etal \cite{simple} trained a fully connected (FC) network to exploit the temporal coherence from the 3D/2D pose sequence. Similarly, Hossain \etal~\cite{hossain2018exploiting} employed LSTM to predict 3D poses from 2D poses in a sequence-to-sequence manner. Considering the computation efficiency of RNN-based methods is limited, temporal convolution network (TCN)~\cite{tcn} is proposed and widely used for temporal modeling. For example, Pavllo \etal \cite{3dvideopose} employed the TCN to predict 3D pose from a consecutive 2D pose sequence. They also conduct back-projection of estimated poses from 3D to 2D for supervision using the labels generated by the state-of-the-art 2D pose detector. Currently, as we state before, TCN-based methods still lack temporal supervision. We propose an adversarial training strategy to use the temporal order as supervision.


\textbf{Recovering 3D mesh from monocular video.} 
Few methods are proposed to recover human 3D mesh from monocular video. Generally, we can split the existing methods into two categories: optimization-based and CNN-based. For example, SFV~\cite{peng2018sfv} first predicts human 3D meshes of each frame by CNN and then smooths the single-frame results via traditional optimization-based post-processing. However, compared with CNN, the efficiency of optimization-based methods is limited. Most recently, HMR-video \cite{kanazawa2018learning} employed the TCN to exploit the temporal information. They proposed to learn the motion dynamic by predicting the actions before and after the current frame. The proposed method outperforms their methods on both 3DPW and Human3.6M.


\section{Method}
\label{sec:method}
The proposed method is to estimate the 3D human mesh parameters $\boldsymbol{\Theta}$ from single images and monocular videos. 
As shown in \textit{Figure}~\ref{fig:flowchart}, the proposed framework has two parts: a) spatially, extracting features of each frame using the proposed DSD module (red) and b) temporally, learning the short and long-term temporal coherence of adjacent frames. 

For single images, the most straightforward method is to estimate $\boldsymbol{\Theta}$ directly from features extracted using Resnet-50~\cite{resnet}. We follow this simple pipeline and insert a DSD module before the final fully connected layer. The proposed DSD module can re-organize the coupled features and disentangle skeleton from the rest details in feature space. 

For monocular videos, we propose a temporal network (SATN) to estimate $\boldsymbol{\Theta}$ from a tuple of features (light green) re-organized by DSD module. More specifically, from $n$ ($n=5$ in \textit{Figure}~\ref{fig:flowchart}) frame features centered at frame $t$, SATN is built to predict the $\boldsymbol{\Theta_t}$. SATN consists of self-attention module (blue) and TCN (gray). 

Besides, an adversarial training strategy (gold) is designed to help the self-attention module learning motion dynamics in the video. In the middle of \textit{Figure}~\ref{fig:flowchart}), features of adjacent frames are shuffled. Self-attention module along with the sequence sorting module is trained to recover the correct temporal order. This strategy is performed in a multi-task learning manner. The correct order of frame features is recovered after the self-attention module. Features in correct temporal order are sent to TCN for predicting the human 3D mesh $\boldsymbol{\Theta_t}$ as usual. 

\subsection{3D Human Body Representation} \label{SMPL}

A parametric statistical 3D human body model, SMPL, is employed to encode the 3D mesh into low-dimensional parameters. SMPL disentangles the shape and pose of a human body. It establishes an efficient mapping $M( \boldsymbol{\beta}, \boldsymbol{\theta};\Phi):\mathbb{R}^{| \boldsymbol{\theta}|\times| \boldsymbol{\beta}|}\mapsto\mathbb{R}^{3 \times 6890}$ from shape $  \boldsymbol{\beta}$ and pose $ \boldsymbol{\theta}$ to a triangulated mesh with 6890 vertices, where $\Phi$ represents the statistical prior of human body. The shape parameter $ \boldsymbol{\beta} \in \mathbb{R}^{10}$ is the linear combination weights of 10 basic shape. The pose parameter $ \boldsymbol{\theta} \in \mathbb{R}^{3 \times 23}$ represents relative 3D rotation of 23 joints in axis-angle representation. To unify the pose formats of different datasets, 14 common joints of LSP~\cite{lsp} are selected. A linear regressor $P_{3d}$ is developed to derive these 14 joints from 6890 vertices of human body mesh. The linear combination operation of this regressor guarantees that joints location is differentiable with respect to shape $\boldsymbol{\beta}$ and pose $\boldsymbol{\theta} $ parameters. 

A weak-perspective camera model is employed in this task to establish the mapping from 3D space to 2D image plane, in convenience of supervising 3D mesh with 2D pose labels. Finally, a 85 dimensional vector $\boldsymbol{\Theta} = \{ \boldsymbol{\theta}, \boldsymbol{\beta}, \boldsymbol{R},\boldsymbol{t},\boldsymbol{s}\}$ is adopted to represent a 3D human body in camera coordinate, where $ \boldsymbol{R} \in \mathbb{R}^{3}$ is the global rotation in axis-angle representation, $ \boldsymbol{t} \in \mathbb{R}^{2}$ and $ \boldsymbol{s} \in \mathbb{R}$ represents translation and scale in image plane, respectively. The projection of $M( \boldsymbol{\beta}, \boldsymbol{\theta};\Phi)$ is
\begin{equation}
	\begin{aligned}
		\boldsymbol{x} = \boldsymbol{s} \Pi(\boldsymbol{R}M( \boldsymbol{\beta}, \boldsymbol{\theta};\Phi)) + \boldsymbol{t},
	\end{aligned}
\end{equation}
where $\Pi$ is an orthographic projection.

\subsection{DSD Module\label{dsd}}
 First of all, we need to extract the 3D mesh information from each frame. As we introduced before, the existing single-frame methods suffer from severe coupling problem. To overcome this problem, we propose the DSD module to disentangle the skeleton from the rest part of the human 3D mesh, including body 3D shape and the detailed poses like the orientation of head, hands, and feet. The DSD module brings two main advantages: a) decoupling reduces the complexity by meticulously task dividing; and b) skeleton-disentangled representation is a better foundation for exploring the temporal motion dynamics. 

As shown in the red part of \textit{Figure}~\ref{fig:flowchart}, a two-branch structure is designed to extract the skeleton and the rest detailed features separately. We follow~\cite{integral} to estimate joint coordinates of the 2D/3D skeleton with minor modifications. Three de-convolution layers are stacked on top of the backbone and followed by DIR (differential integral regression) to convert the normalized heatmaps $H$ into 2D/3D joint coordinates. The $DIR(H)$ is the integration of all location indices $p$ in the heatmaps, weighted by their probabilities. For instance, the 3D coordinate values of the $k$-th joint $J_k$ could be derived from the $k$-th 3D heatmap $H_k$ by
\begin{equation}\label{equ:integral}
 \begin{aligned}
J_k = \sum_{p_z=1}^{D} \sum_{p_y=1}^{H} \sum_{p_x=1}^{W} pH_k(p),
 \end{aligned}
\end{equation}
where $D$,  $H$, and $W$ are depth, height and width of the $k$-th heatmap $H_k$ respectively. For 2D heatmaps, $D=1$.

After independent feature extraction, we need to find a proper way to aggregate them. For the two-factor problems, bilinear transformation is well-known for its strong ability of decoupling the style-content factors~\cite{bilinearfreeman}, such as the identity and head orientation in face recognition; or accent and word class in speech recognition. Therefore, we employ it to disentangle the skeleton from the rest details.
Given the 2D skeleton coordinates $\boldsymbol{x_s} \in \mathbb{R}^{N\times 28} $ and corresponding detailed features $\boldsymbol{x_d} \in \mathbb{R}^{N \times 512}$ , their bilinear transformation $\boldsymbol{y} \in \mathbb{R}^{N \times 512}$ is 
\begin{equation}\label{equ:bilinear}
 \begin{aligned}
\boldsymbol{y} = \boldsymbol{x_s}\boldsymbol{A}{\boldsymbol{x_d}}^T,
 \end{aligned}
\end{equation}
where $\boldsymbol{A} \in \mathbb{R}^{512 \times 28 \times 512}$ is the learnable weights. 

Moreover, with high modularity and the simple design, it is convenient to transplant DSD module to other networks. For instance, DSD module could be easily plugged into the existing 2D/3D pose estimation network for predicting 3D human body meshes. As shown in \textit{Figure}~\ref{fig:flowchart}, two fully connected layers, and a bilinear transformation layer are all we need for this conversion. 

\subsection{Self-attention Temporal Network}
Secondly, we propose a self-attention temporal network (SATN) for learning long and short-term temporal coherence in the video. Given skeleton-disentangled features from DSD module, SATN predicts smoother $\Theta$ in temporal. TCN performs well in modeling the short-term patterns and efficient parallel computation. But due to its inherent hierarchical structure, it is inefficient to model the long-term correlation in the video. To solve this problem, we propose to build the TCN on top of the self-attention module. As illustrated in \textit{Figure}~\ref{fig:flowchart}, given a tuple of frame features $\boldsymbol{x_f} \in \mathbb{R}^{N \times 512}$ centered at $t$ , self-attention module is employed to relate different frames and establish the long-term sequence representations. Then, the outputs of the self-attention module are fed into the TCN for predicting the human 3D mesh $\boldsymbol{\Theta_t}$. 

In particular, the positional encoding is added to each input feature vector for injecting information about their absolute position in sequence. Multi-head attention (MHA)~\cite{transformer} is adopted to accomplish the self-attention mechanism, which is a variance of typical scaled dot-product attention (SDPA)~\cite{transformer}. In this work, we employ 8 heads for MHA. In detail, $\boldsymbol{x_f}$ is first linearly projected to $\{\boldsymbol{x_f^i} \in \mathbb{R}^{N \times 64}, i=1,...,8\}$ via 8 fully connected (FC) layers. Then SDPA is performed on each $\{\boldsymbol{x_f^i}\}$ in parallel. The outputs are concatenated and linearly projected to $\boldsymbol{y_f} \in \mathbb{R}^{N \times 512}$ using a FC layer. The SDPA is derived as  
\begin{equation}\label{equ:sdpa}
 \begin{aligned}
SDPA(\boldsymbol{x_f}) = softmax(\frac{\boldsymbol{x_f} \boldsymbol{x_f}^T}{\sqrt{d}})\boldsymbol{x_f},
 \end{aligned}
\end{equation}
where d=512 is the dimension of inputs and serves as the scaling factor. In \textit{Equation}~\ref{equ:sdpa}, the output is the weight sum of $\boldsymbol{x_f}$. The weight matrix represents the relation between each two frames. The paths length between different frames is reduced to constants. With the assistance of the self-attention module, the bottom layers of TCN could reach to the information of entire sequence. The receptive field is relatively expanded. Short and long-term temporal coherence can be learned more efficiently.

Although we predict from multi-frame features in temporal, the supervision is still in single-frame level. We lack the temporal supervision for the long-term representation learning. To tackle this problem, we propose an unsupervised adversarial training strategy. Similar to recent methods~\cite{misra2016shufflebinary,sumer2017shufflebinary,fernando2017shufflebinary,sortingsequence}, we also use the temporal order of frames as the supervision for the representation learning. However, there are some differences in the formulation of the problem. Most of the previous methods~\cite{misra2016shufflebinary,sumer2017shufflebinary,fernando2017shufflebinary} formulated the problem as the binary classification that verifies the correct/incorrect temporal order. Lee \etal~\cite{sortingsequence} developed it to predict $n!/2$ combinations for each n-tuple of frames. In this work, we further develop it to directly estimate the original position indices of the shuffled frames for richer supervision. Besides, two adaptive changes of loss function are made to meet the special properties of the motion sequence.


As illustrated in the \textit{Figure}~\ref{fig:flowchart}, frame features are first shuffled before input. As before, the input is added up with the positional coding, which stays unchanged to avoid disclosure of the order information. After going through the self-attention module, the outputs are sent to the TCN and the sequence sorting module separately. Before fed into TCN, the shuffled order is retrieved to avoid the interference with the normal representation learning of the TCN. The sequence sorting module predicts the position indices of the correct order. Specifically, for each tuple of 9 shuffled frames, we need to predict 9 indices, indicating their positions in the original sequence. Besides, two adaptive modifications of sorting target are made. Firstly, considering the reasonable orders of some actions (e.g., standing up/sitting down) are reversible, the minor loss between the predicted and the forward/backward ground truth orders will be chosen. Secondly, sometimes, the differences between adjacent frames are too small to be properly preserved after single-frame convolution encoding. In this situation, the context order of adjacent frames becomes ambiguous. But the typical one-hot label will make equal punishments on this kind of failures without considering the adjacent ambiguity. To tackle this problem, we replace the hard one-hot label with the soft Gaussian-like label for proper supervision. 

\subsection{Loss Functions}

As we mentioned in \textit{Section}~\ref{sec:intro}, full 3D annotations of 2D image available are limited to the experimental environment. Models trained on these data cannot generalize well to the in-the-wild images.  To make full use of existing 2D/3D data, the estimated 3D human body parameters are optimized with 
\begin{equation}\label{equ:ltheta}
 \begin{aligned}
 L_{\boldsymbol{\Theta}} & = w_{pm}L_{pm} + w_{3d}L_{3d} + w_{2d}L_{2d} + w_{r}L_{r}
 \end{aligned}
\end{equation}
where $w_{pm}, w_{3d}, w_{2d}, w_{r}$ are weights of these loss items. For images with motion capture, $L_{pm}$ is employed to supervise the the body parameter $\boldsymbol{\theta}$, acquired by moshing~\cite{mosh}, with L2 loss directly. In particular, inspired by~\cite{hmr,humanshape}, angular pose $\boldsymbol{\theta} \in \boldsymbol{\Theta}$ is converted from rotation vectors to $3\times 3$ rotation matrices via the Rodrigues formula for stable training. For images with 2D/3D pose annotations, $L_{2d}$ and $L_{3d}$ are employed to supervise the skeleton pose $J^{2d}$/$J^{3d}$ of estimated body mesh using L1/L2 loss respectively. $J^{3d}$ and $J^{2d}$ are derived from the predicted $\boldsymbol{\Theta} = \{ \boldsymbol{\theta}, \boldsymbol{\beta}, \boldsymbol{R},\boldsymbol{t},\boldsymbol{s}\}$ by
\begin{equation}\label{equ:j3}
 \begin{aligned}
 J^{3d} & = P_{3d}(M( \boldsymbol{\beta}, \boldsymbol{\theta};\Phi))
 \end{aligned}
\end{equation}
and
\begin{equation}\label{equ:j2}
 \begin{aligned}
 J^{2d} & = \boldsymbol{s}\Pi(\boldsymbol{R}J^{3d})+\boldsymbol{t},
 \end{aligned}
\end{equation}
where $M$ represents a mapping from shape $\boldsymbol{\beta}$ and pose $\boldsymbol{\theta}$ to a 3D human body mesh, $P_{3d}$ represents a linear mapping from a 3D mesh to 3D joints coordinates and $\Pi$ is an orthographic projection, as we defined in \textit{Section}~\ref{SMPL}. 

Besides, it is important to establish rational constraints of joint angles and body shape, especially for learning from images with only 2D pose annotations. Therefore, we follow~\cite{hmr} and employ a discriminator to provide rationality loss $L_{r}$ using Mocap data~\cite{mocap}. Besides, $L_{2DJ}$ is employed to supervise the 2D skeleton coordinates $x_s$ in the DSD module using L1 loss. The sorting results of sequence sorting module are supervised with L2 loss of $L_{s}$.

\section{Experiments}\label{sec:experiments}

\begin{figure}
	\begin{center}
		\includegraphics[width=1\linewidth]{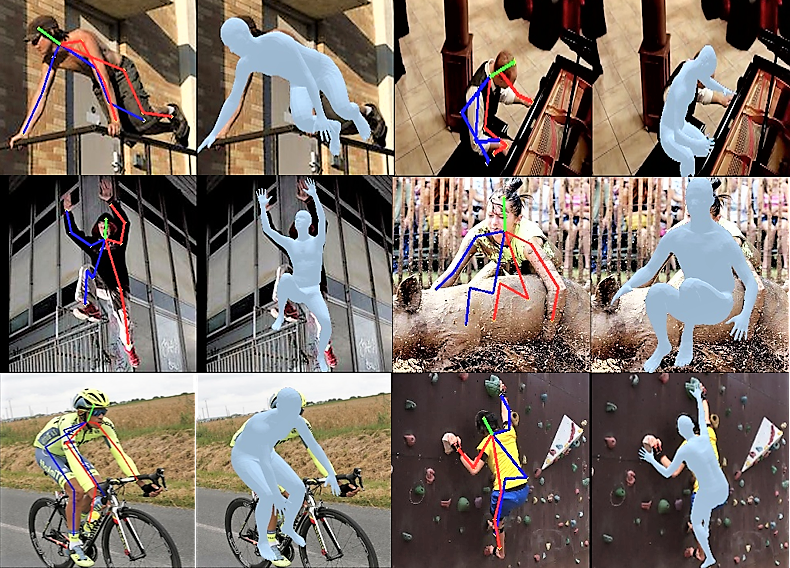}
		\caption{Some qualitative results of DSD network.}
		\label{fig:dsdresult}
	\end{center}
\end{figure}

\subsection{Datasets} \label{Dataset}

\noindent\textbf{Human3.6M} is the only dataset with 3D annotations we used to train. It contains videos of multiple actors performing 17 activities, which are captured in a controlled environment. We downsample all videos from 50fps to 10fps for removing the redundancy. The proposed method is evaluated on common protocol following~\cite{hmr} for a more comprehensive comparison. In details, we train on 5 subjects (S1, S5, S6, S7, S8) and test on subject S9 and S11. We report both the mean per joint position error (MPJPE) and Procrustes Aligned MPJPE (PA-MPJPE), which is MPJPE after rigid alignment of predicted pose with ground truth, in millimeters. Except for the common MPJPE and PA-MPJPE, mean per joint velocity and acceleration error (MPJVE/MPJAE), in $mm/s$ and $mm/s^2$ respectively, are adopted to evaluate the smoothness and stability of predictions over time. 

\noindent\textbf{3DPW}~\cite{3dpw} is a recent challenge dataset that contains 60 video sequences (24 train, 24 test, 12 validation) of richer activities, such as climbing, golfing, relaxing on the beach, etc. They leverage video and IMU to obtain accurate 3D pose despite the complexity of scenes. For a fair comparison, none of the approaches get trained on 3DPW. We report PA-MPJPE on all splits (train/test/val) of this dataset.

\noindent\textbf{2D in-the-wild datasets.} We use MPII~\cite{mpii}, LSP~\cite{lsp}, AICH~\cite{aich}, and Penn Action~\cite{pennaction}, which are 2D pose datasets without 3D annotations, to train our single-frame DSD network for better generalization. We follow~\cite{keep,unite,hmr} to use 14 LSP joints as the skeleton.

\subsection{Implement Details}
\textbf{Architecture:} The framework has spatial and temporal parts. In spatial, features are extracted using ResNet-50, pre-trained on MPII, and further decoupled using DSD module. In temporal, features of video clips are gathered and put into SATN, which compose of self-attention module and TCN. The self-attention module contains 2 Transformer~\cite{transformer} blocks. TCN follows the design of~\cite{3dvideopose,tcn} but only contains 2 convolution blocks. Since the Human3.6M is down-sampled, we take a receptive filed of 9 frames for the TCN, equivalent to 45 frames in the original video. The sequence sorting module consists of 3 convolution blocks followed by 3 FC layers. 

\textbf{Training details:} We train both single-frame DSD network and SATN for 40 epochs. Considering the efficiency of training, we pre-compute the single-frame features and train SATN with them directly. The SGD is adopted as the optimizer with momentum=0.9. The learning rate and batch size are set to 1e-4 and 16 respectively. The weights of loss items are set as $w_{pm}=20, w_{r}=0.6, w_{2d}=10, w_{3d}=60$. Besides, considering the domain gap between the train and the test set, we discard the hyper-parameters of the batch normalization layers during the evaluation. Besides, to enhance the stability of occlusion, we enlarge the scale augmentation to generate some samples with half body.

\begin{table}
  \begin{center}
  \begin{tabular}{lccc}
  \hline
  Method & Train & Test & val\\ 
  \hline
  Simple-baseline~\cite{simple} & - & 157.0 & -\\
  SMPLify~\cite{keep} & - & 106.8 & -\\
  TP-Net~\cite{tpnet} & -  & 92.2 & -\\
  HMR-video~\cite{kanazawa2018learning} & - & 80.1 & -\\
  HMR-video-L~\cite{kanazawa2018learning} & 75.9 & 72.6 & 76.7\\
  \textbf{DSD} & 75.0  & 75.0 & 78.0\\
  \textbf{DSD+SATN} & \textbf{68.3} & \textbf{69.5} & \textbf{71.8}\\
  \hline
  \end{tabular}
  \end{center}
  \caption{Comparisons to state-of-the-art methods on entire 3DPW in terms of PA-MPJPE without any fine-tuning.}
  \label{tab:comp3dpw}
\end{table}

\subsection{Comparisons to State-of-the-art Approaches}
The results in \textit{Table}~\ref{tab:comp3dpw} shows that the proposed methods achieve the state-of-the-art results on 3DPW. Again, all approaches are directly tested without any fine-tuning and Human3.6M is the only 3D train set. It demonstrates that the proposed methods perform well in generalization. The most similar method to ours is HMR-video~\cite{kanazawa2018learning}, which also learns motion dynamic with TCN for temporal optimization. We compared with their two evaluation settings, HMR-video and HMR-video-L. The main difference between two settings is that HMR-video is trained with datasets of similar scale with ours, while HMR-video-L is trained with their internet dataset which is nearly 20x larger than ours. Our approach outperforms HMR-video in terms of PA-MPJPE by 13.2\% on 3DPW test set. On all splits of 3DPW, we evaluate HMR-video-L model they released. As shown in \textit{Table}~\ref{tab:comp3dpw}, the proposed method superior to HMR-video-L on all splits. 

State-of-the-art results are also achieved on Human3.6M. In \textit{Table}~\ref{tab:h36mcomp1}, the proposed method outperforms the HMR-video and HMR-video-L by 26.6\% and 23.8\% in terms of PA-MPJPE respectively. Note that, in \textit{Table}~\ref{tab:comp3dpw}, TP-Net and Simple-baseline, also trained on Human3.6M, perform extremely well on Human3.6M, while showing poor generalization ability on 3DPW. Besides, the proposed method greatly outperforms SMPLify on 3DPW, which indicates that the improvements are not brought by the different 3D pose representation.

\begin{table}
  \begin{center}
  \begin{tabular}{lcc}
  \hline
  Method & MPJPE$\downarrow$ &  PA-MPJPE$\downarrow$  \\
  \hline
  HMR~\cite{hmr} & 87.9 & 58.1\\
  STN~\cite{yoshiyasu2018skeleton} & 69.9 & 61.4\\
  HMR-video & - & 57.8\\
  HMR-video-L & - & 55.7\\
  \textit{Direct} & 110.0 & 62.8\\
  \textit{Concat} & 91.8 & 63.1\\
  \textbf{DSD} & 60.1 & 44.3 \\
  \textbf{DSD+SATN} & \textbf{59.1} & \textbf{42.4}\\
  \hline
  \end{tabular}
  \end{center}
  \caption{Comparisons to state-of-the-art methods that predict 3D meshes on Human3.6M. }
  \label{tab:h36mcomp1}
\end{table}

\begin{table}
  \begin{center}
  \begin{tabular}{lcc}
  \hline
  Method & MPJPE$\downarrow$ &  PA-MPJPE$\downarrow$ \\
  \hline
  Backbone+TCN & 87.8 & 58.5\\
  Backbone+SATN & 80.6 & 58.6 \\
  DSD+TCN & 82.3 & 52.0 \\
  DSD+TCN+Self-attention & 59.6 & 43.4 \\ 
  DSD+SATN & \textbf{59.1} & \textbf{42.4} \\
  \hline
  \end{tabular}
  \end{center}
  \caption{Ablation study of the components in SATN.}
  \label{tab:ablationstudy}
\end{table}

\begin{figure}
	\begin{center}
		\includegraphics[width=1\linewidth]{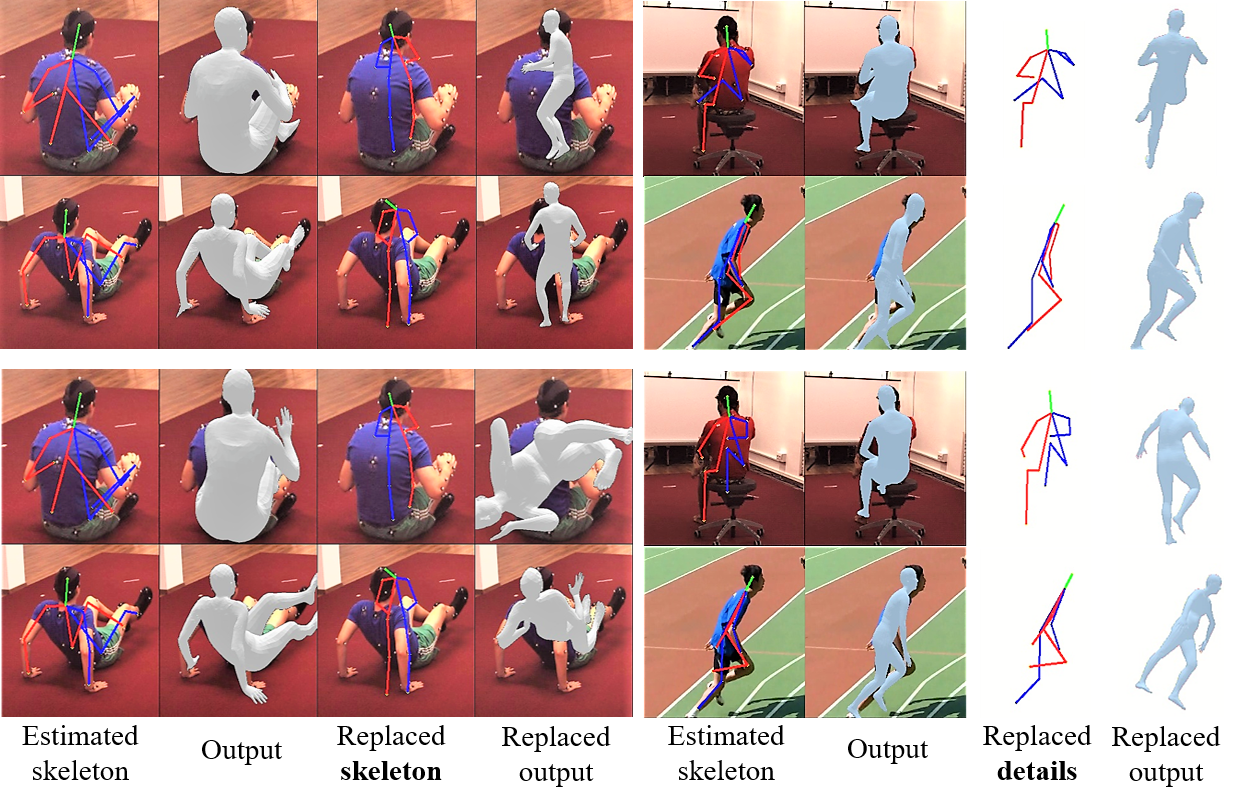}
		\caption{Decoupling effectiveness comparison between DSD module (the two rows upper) and \textit{Concat}(the two rows lower) by replacing the skeleton (shown in the left part) or the detailed features (shown in the right part).}
		\label{fig:decoupleresult}
	\end{center}
\end{figure}

\subsection{Evaluation of the DSD module\label{dsdeval}}

\begin{table*}
  \begin{center}
  \setlength{\tabcolsep}{1mm}{
  \begin{tabular}{lcccccccccccccccc}
  \hline
   Method & Dir. & Disc. & Eat & Greet & Phone & Pose & Purch. & Sit & SitD. & Smoke & Photo & Wait & Walk & WalkD. & WalkT. & Avg.$\downarrow$ \\
  \hline
\textit{Direct}&14.5&15.6&14.9&15.6&15.4&15.0&15.9&15.4&16.1&15.6&15.7&15.7&15.8&16.0&16.1&16.1\\
\textit{Concat}&20.0&20.9&19.4&19.7&19.2&18.8&19.5&19.1&20.1&19.4&19.7&19.6&19.7&20.4&20.4&20.5\\
DSD& 12.3 & 12.8 & 12.2 & 12.6 & 12.4 & 12.1 & 12.1 & 11.8 & 12.3 & 12.0 & 12.0 & 12.0 & 12.2 & 12.4 & 12.5 & 12.5\\
DSD+SATN & 8.8 & 9.1 & 8.5 & 8.9 & 8.7 & 8.6 & 8.7 & 8.3 & 8.4 & 8.3 & 8.3 & 8.3 & 8.4 & 8.6 & 8.7 & 8.7 \\
  \hline
  \end{tabular}}
  \end{center}
  \caption{MPJVE($mm/s$): Velocity error over the poses of the predicted 3D human meshes.}
   \label{tab:MPJVE}
\end{table*}

\begin{table*}
  \begin{center}
  \setlength{\tabcolsep}{1mm}{
  \begin{tabular}{lcccccccccccccccc}
  \hline
   Method & Dir. & Disc. & Eat & Greet & Phone & Pose & Purch. & Sit & SitD. & Smoke & Photo & Wait & Walk & WalkD. & WalkT. & Avg.$\downarrow$ \\
  \hline
\textit{Direct}&13.1&14.2&13.5&14.3&14.0&13.7&14.7&14.0&14.3&14.0&14.0&14.0&14.3&14.6&14.8&14.8\\
\textit{Concat}&17.1&18.0&16.6&17.1&16.5&16.2&16.8&16.2&16.7&16.1&16.4&16.4&16.7&17.3&17.4&17.4\\
DSD & 11.5 & 12.0 & 11.3 & 11.8 & 11.5 & 11.2 & 11.3 & 10.8 & 11.0 & 10.8 & 10.8 & 10.8 & 11.1 & 11.3 & 11.5 & 11.5\\
DSD+SATN & 7.0 & 7.4 & 6.8 & 7.2 & 6.9 & 6.8 & 6.9 & 6.5 & 6.5 & 6.3 & 6.4 & 6.3 & 6.5 & 6.7 & 6.8 & 6.8 \\
  \hline
  \end{tabular}}
  \end{center}
  \caption{MPJAE($mm/s^2$): Acceleration error over the poses of the predicted 3D human meshes.}
   \label{tab:MPJAE}
\end{table*}

\textbf{1) The ablation study of the DSD module.} Two baseline methods are compared in \textit{Table}~\ref{tab:h36mcomp1} for evaluating the DSD module. \textit{Direct} is to directly estimate the $\boldsymbol{\Theta}$ without DSD module. \textit{Concat} is to replace the bilinear transformation layer of DSD module with the concatenation operation. Compared with \textit{Direct}, \textit{Table}~\ref{tab:h36mcomp1} shows that \textit{Concat} performs even worse in PA-MPJPE, but adding DSD module brings significant improvement (by 45.3\% MPJPE and 29.4\% PA-MPJPE). The results demonstrate that decoupling skeleton improves the accuracy of 3D pose recovery. Next, we will validate that the skeleton is sufficiently disentangled from the rest details by the DSD module.

\textbf{2) Decoupling effectiveness.} We set up a visual experiment to compare the decoupling effectiveness of DSD module and \textit{Concat}. For verifying whether the skeleton is sufficiently decoupled, we deliberately replace the estimated skeleton with the randomly selected one. The left part of \textit{Figure}~\ref{fig:decoupleresult} with gray human mesh shows the different responses from DSD module (the upper two rows) and \textit{Concat} (the lower two rows) to this replacement. As we can see, the body postures of 3D mesh predicted by the DSD module are correctly changed together with the replaced skeleton, while the outputs from \textit{Concat} are completely messed up. The experimental results show that the skeleton is successfully disentangled from the rest details. Likewise, a similar conclusion can also be drawn from the right part of \textit{Figure}~\ref{fig:decoupleresult} with blue human mesh, which is the results when we replace the detailed features and keep skeleton unchanged.

\textbf{3) Stabilily.} In addition, \textit{Table}~\ref{tab:MPJVE} and~\ref{tab:MPJAE} show their MPJVE and MPJAE. Note that \textit{Concat} has much larger MPJVE and MPJAE than the rest, even including \textit{Direct}. It indicates that the predictions derived from the concatenated features are more unstable. By contrast, the DSD module improves stability with smoother predictions.

\textbf{4) Comparisons to the state-of-the-art single-frame methods.} In \textit{Table}~\ref{tab:h36mcomp1}, we compare proposed DSD module with HMR and STN. We tightly follow the same evaluation protocol and use the same backbone, ResNet-50, as stated in their articles. Compared with HMR, adding DSD module reduces the error by 31.6\% MPJPE and 23.7\% PA-MPJPE, which further proves the effectiveness of DSD module. We also compare to STN~\cite{yoshiyasu2018skeleton}, which is the most recent proposed method that also predicts angular pose and outputs 3D meshes. In \textit{Table}~\ref{tab:h36mcomp1}, DSD network outperforms the STN by 14\% MPJPE and 27.9\% PA-MPJPE. Some qualitative results of DSD module are illustrated in Figure~\ref{fig:dsdresult}.

\subsection{Evaluation of SATN}

\textbf{1) The ablation study of SATN.} \textit{Table}~\ref{tab:h36mcomp1} shows that compared with the single-frame DSD network, adding SATN brings an additional 4.2\%/7.3\% improvement in terms of MPJPE/PA-MPJPE on Human3.6M. In addition, \textit{Table}~\ref{tab:comp3dpw} shows that adding SATN brings a higher improvement (7.3\% PA-MPJPE) on 3DPW. Besides, except for reducing the error of 3D pose recovery, as shown in \textit{Table}~\ref{tab:MPJVE} and~\ref{tab:MPJAE}, SATN significantly improves the smoothness of the predictions. 

\textbf{2) The ablation study of components in SATN.} \textit{Table}~\ref{tab:ablationstudy} shows that performance gets steadily improved by adding the proposed components, including the self-attention module and the adversarial training strategy. Note that making temporal modeling simply with the TCN is not enough. Compared with the results of the single-frame DSD network in \textit{Table}~\ref{tab:h36mcomp1}, \textit{Table}~\ref{tab:ablationstudy} shows that DSD+TCN leads to the degradation of performance. Similar conclusions can also be drawn from HMR-video~\cite{kanazawa2018learning}. As shown in \textit{Table}~\ref{tab:ablationstudy}, involving self-attention module greatly improves the performance of the temporal network and reverses the degradation. This phenomenon demonstrates that it is vital for TCN to establish associations of the entire input sequence as early as possible and self-attention module performs well in this task. 

\subsection{Role of the DSD Module for SATN} 
The results in \textit{Table}~\ref{tab:ablationstudy} shows that skeleton-disentangled representation is of great importance for effective temporal modeling. Backbone+TCN in \textit{Table}~\ref{tab:ablationstudy} is to train the TCN with features from the backbone directly without DSD module. As shown in \textit{Table}~\ref{tab:h36mcomp1}, the results of Backbone+TCN are comparative to the HMR-video~\cite{kanazawa2018learning}, which has similar architecture. By contrast, involving DSD module (DSD+TCN in \textit{Table}~\ref{tab:ablationstudy}) brings a 11.1\% improvement in terms of PA-MPJPE. Besides, when we replace TCN with SATN, the performances get further improvement. Especially, in \textit{Table}~\ref{tab:ablationstudy}, the combination of DSD and SATN  (DSD+TCN+Self-attention+adversarial training) causes a magic reaction and outperforms all backbone-based methods by 26.6\% MPJPE and 27.5\% PA-MPJPE. However, PA-MPJPE of Backbone+TCN and Backbone+SATN in \textit{Table}~\ref{tab:ablationstudy} are nearly equal, which indicates that skeleton-disentangled representation is essential for motion modeling. 

\section{Conclusion}
We propose an end-to-end framework for recovering 3D human mesh from single images and monocular videos via a skeleton-disentangled representation. Decoupling skeleton reduces the complexity and the 3D pose error. The proposed DSD module could be regarded as an efficient bridge between 2D/3D pose estimation and 3D mesh recovery. Moreover, SATN is well designed to explore long and short-term temporal coherence. Massive evaluations demonstrate that we provide an attractive and efficient baseline method for related problems, including human motion analysis, 3D virtual try-on and multi-person 3D meshes recovery. 

{\small
\bibliographystyle{ieee_fullname}
\bibliography{egpaper_final}
}

\end{document}